\documentclass[letterpaper]{article} 
\usepackage{aaai25}  
\usepackage{times}  
\usepackage{helvet}  
\usepackage{courier}  
\usepackage[hyphens]{url}  
\usepackage{graphicx} 
\usepackage{natbib}  
\usepackage{caption} 
\usepackage{algorithm}
\usepackage{algorithmic}
\usepackage{cuted}
\usepackage{amsmath}
\usepackage{pifont}
\usepackage{amsfonts}
\usepackage{multirow}
\usepackage{makecell}

\usepackage{newfloat}
\usepackage{listings}
\DeclareCaptionStyle{ruled}{labelfont=normalfont,labelsep=colon,strut=off} 
\floatstyle{ruled}
\newfloat{listing}{tb}{lst}{}
\floatname{listing}{Listing}
\title{Video Diffusion Models are Strong Video Inpainter}
\author{
    Minhyeok Lee, 
    Suhwan Cho, 
    Chajin Shin, 
    Jungho Lee, 
    Sunghun Yang, 
    Sangyoun Lee
}
\affiliations{
    Yonsei University\\
    Seoul, Republic of Korea\\
    \{hydragon516,chosuhwan,chajin,2015142131,sunghun98,syleee\}@yonsei.ac.kr
}

\usepackage{bibentry}

\begin{document}

\maketitle

\begin{strip}
	\vspace{-63pt}
	\centering
	\includegraphics[width=1.0\textwidth]{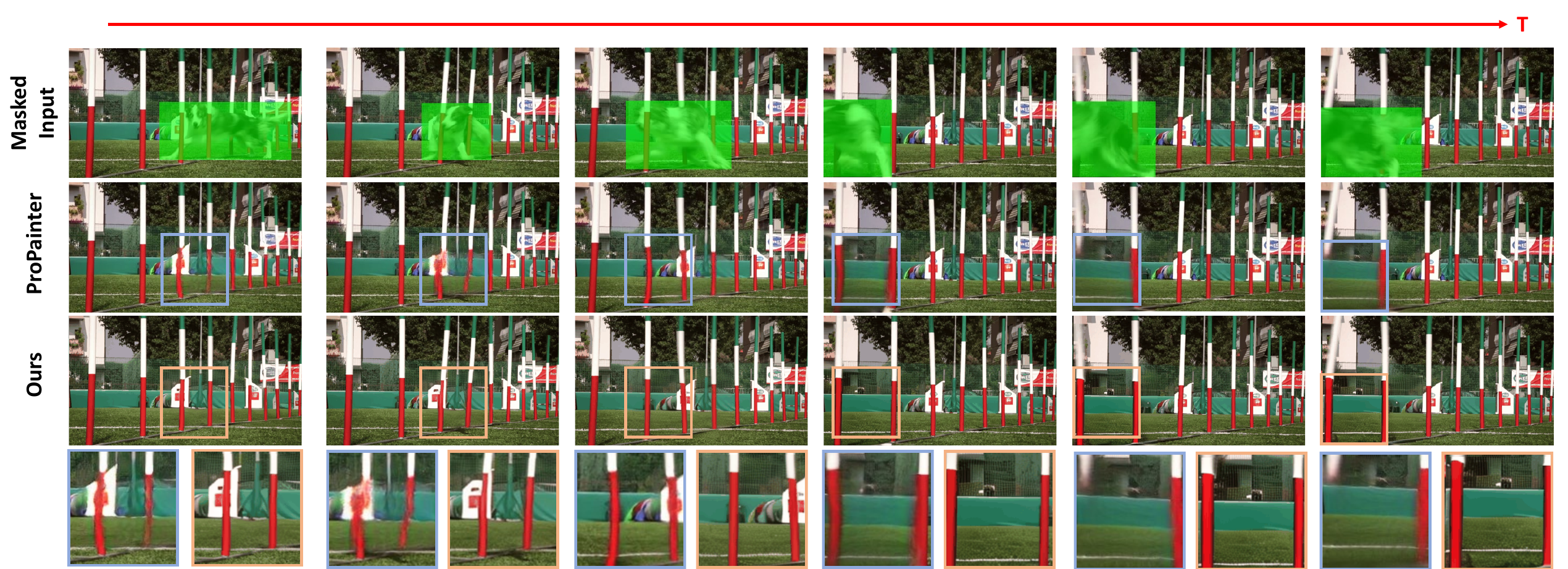}
	\captionof{figure}{
		Inpainting results of the proposed FFF-VDI and the flow propagation-based ProPainter. When the target object is frequently occluded or structurally difficult to track, large and rough bounding box masks are advantageous for editing.}
	\label{fig:intro}
\end{strip}

\begin{abstract}
	Propagation-based video inpainting using optical flow at the pixel or feature level has recently garnered significant attention. However, it has limitations such as the inaccuracy of optical flow prediction and the propagation of noise over time. These issues result in non-uniform noise and time consistency problems throughout the video, which are particularly pronounced when the removed area is large and involves substantial movement. To address these issues, we propose a novel First Frame Filling Video Diffusion Inpainting model (FFF-VDI). We design FFF-VDI inspired by the capabilities of pre-trained image-to-video diffusion models that can transform the first frame image into a highly natural video. To apply this to the video inpainting task, we propagate the noise latent information of future frames to fill the masked areas of the first frame's noise latent code. Next, we fine-tune the pre-trained image-to-video diffusion model to generate the inpainted video. The proposed model addresses the limitations of existing methods that rely on optical flow quality, producing much more natural and temporally consistent videos. This proposed approach is the first to effectively integrate image-to-video diffusion models into video inpainting tasks. Through various comparative experiments, we demonstrate that the proposed model can robustly handle diverse inpainting types with high quality.
\end{abstract}

\section{Introduction}
\label{intro}
Video inpainting (VI) aims to fill in missing regions of a video with visually consistent content while ensuring both spatial and temporal consistency. Unlike image inpainting, video inpainting faces the significant challenge of establishing accurate correspondences with distant frames to obtain missing pixel information and maintaining time consistency between video frames to generate natural videos. To address these issues, recent methods~\cite{gao2020flow, zhang2022inertia, zhang2022flow, zhou2023propainter} based on information propagation through optical flow have gained significant attention. These models use inpainted optical flow information to propagate pixel or feature information to the masked areas of each frame. While this approach can generate relatively accurate images by propagating actual pixel information to the missing regions, it has several significant drawbacks. Firstly, the quality of the predicted video is highly dependent on the accuracy of the completed optical flow, since pixel information is propagated based on the optical flow between all adjacent frames. Incorrect predictions can accumulate error noise during the frame propagation process, reducing the overall quality of the video frames. Secondly, in the case of object removal, propagation-based models require relatively accurate target object masks. These methods are highly dependent on the amount of visual cues from the unmasked areas of the reference frames to fill the masked regions of the target frame. Therefore, to ensure consistent performance, the mask size must be minimized, which necessitates precise target object masks. However, this leads to increased manual frame-by-frame masking costs in real-world applications.

Recently, a few works~\cite{zhang2024avid, zi2024cococo, gu2023flow} integrate the powerful generative capabilities of image or video diffusion models into video inpainting tasks. These methods are expected to be more robust to mask size and generate perceptually improved videos compared to propagation-based methods. However, despite the advancements in diffusion models, there are two main reasons that make it challenging to apply them to video inpainting. First, diffusion models do not consider the objects behind the moving mask areas. As the mask's position shifts over time, the actual pixel information in the previously masked areas is revealed. If the past frames generated by the diffusion model differ from the real context, this leads to videos with awkward consistency over time. This becomes even more unnatural when new objects hidden behind the mask appear. Second, when using pre-trained video diffusion models, object hallucination often occurs, where unwanted objects are generated in the mask area. This phenomenon is due to the diffusion model's powerful video generation capabilities, and to prevent it, text guidance is required for the diffusion model.

To address these issues, we propose a new First Frame Filling Video Diffusion Inpainting model, named FFF-VDI. We design FFF-VDI, inspired by the capabilities of pre-trained image-to-video diffusion models to generate complete videos based on the first frame image. Our model has the following advantages: First, the proposed model extracts the latent code of each frame using a VAE encoder to create the masked noise latent code. Next, we use optical flow to propagate the noise latent codes of future frames to fill the missing parts of the first frame's noise latent code. Finally, we fill the masked areas of the future frames with random noise and use the Deformable Noise Alignment (DNA) module to improve temporal consistency and minimize distortion at the noise latent level. By completely filling the first frame latent information with information from future frames and using these as conditional features, we can fine-tune the pre-trained image-to-video diffusion model to generate much more natural inpainting videos. 

This approach has the advantage of significantly reducing dependency on the accuracy of completed optical flow compared to traditional methods~\cite{gao2020flow, zhang2022inertia, zhang2022flow, zhou2023propainter} that propagate all frame pixels or feature information using optical flow. Figure~\ref{fig:intro} shows the object removal performance of ProPainter~\cite{zhou2023propainter}, which is based on optical flow propagation, and the proposed FFF-VDI. As shown in the figure, the existing method applies optical flow propagation to all frames, resulting in errors accumulating and causing texture blurring or distortion in the resulting video. However, the proposed method applies noise latent propagation only to the first frame and utilizes the strong temporal consistency performance of the video diffusion model for the subsequent frames, resulting in much more consistent and natural videos. Furthermore, our method considers the actual pixel information of the areas occluded by the mask. Traditional video diffusion models prioritize maintaining temporal consistency between neighboring frames. However, the proposed method brings in future latent information to reconstruct the actual pixel information in the masked areas of the first frame. As a result, temporal consistency is maintained even when the mask moves or previously occluded areas are revealed. Furthermore, to address the hallucination effects inherent in traditional diffusion models without using text guidance, the proposed FFF-VDI inference process applies DDIM inversion~\cite{mokady2023null}. We conduct DDIM inversion~\cite{mokady2023null} to the entire video frame, propagating the inverted noise to the first frame. This inverted noise generates only the actual erased areas during the denoising process, thus minimizing object hallucination compared to traditional methods that use random noise.

We conduct various comparative experiments to demonstrate that the proposed FFF-VDI outperforms previous methods in both video completion and object removal across different scenarios. Furthermore, we show that our method has robust performance with large and rough masks for object removal, compared to existing optical flow propagation-based methods.

\begin{figure*}[t]
	\centering
	\includegraphics[width=\linewidth]{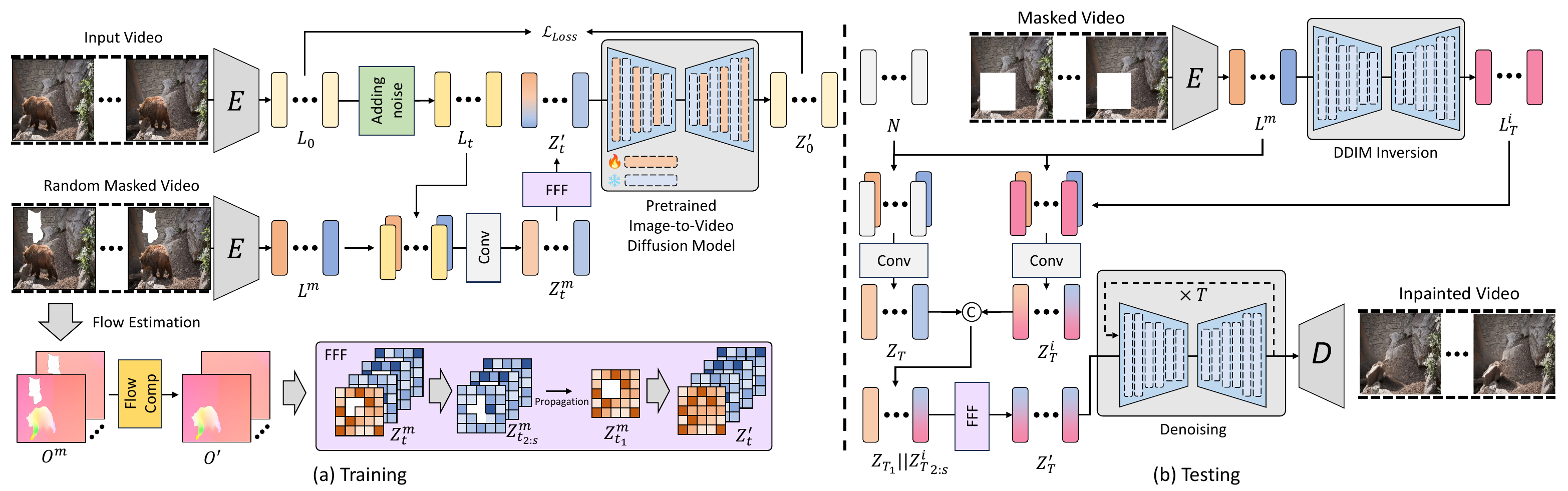}
	\caption{The overall training and testing pipeline structure of our FFF-VDI.}
	\label{fig:main}
\end{figure*}

\section{Related Work}
\textbf{Flow propagation-based approaches.} Flow propagation-based approaches leverage the relatively simpler task of flow completion to assist in the more complex task of RGB content filling. Flow-based video inpainting methods generally follow a three-phase process: flow completion, content propagation, and content generation. Various methods have been proposed to improve each phase. For example, FGVC~\cite{gao2020flow} integrates gradient propagation during the content propagation phase, and E2FGVI~\cite{li2022towards} introduces an end-to-end flow completion module. Additionally, FGT~\cite{zhang2022flow} combines decoupled spatiotemporal attention with gradient propagation techniques from FGVC. ProPainter~\cite{zhou2023propainter} pushes the boundaries further by integrating dual-domain propagation. However, these methods often struggle with spatial misalignment due to flow propagation errors or suffer from diminished detail retention caused by repetitive re-sampling in attempts to achieve sub-pixel accuracy.

\noindent
\textbf{Diffusion-based approaches.} Diffusion-based methods generate inpainted frames by leveraging the generative capabilities of the diffusion model and a module that maintains temporal consistency at the noise level. For example, AVID~\cite{zhang2024avid} uses an image diffusion model to naturally fill each frame and applies a temporal consistency module during the inference stage to generate inpainted videos. CoCoCo~\cite{zi2024cococo} is similar to AVID but additionally introduces a motion capture module to maintain motion consistency in the generated video. However, these methods focus on improving perceptual temporal consistency and do not consider the actual pixel information in the masked areas.

\section{Preliminaries}
\textbf{Video Diffusion Model.} Most video-based diffusion models~\cite{rombach2022high, blattmann2023stable, ho2022video} use a 3D-Unet to learn how to remove noise sequences randomly sampled from a Gaussian distribution. First, a pre-trained 2D-VAE~\cite{kingma2013auto} encoder $E\left(\cdot \right)$ is used to extract the latent codes for each of the $S$ frames of the input video, forming a video latent code sequence $\left \{ Z _ { 0 _ { i } } \right \} _ { i=1 } ^ { S }$. Then, this latent sequence is concatenated along the temporal dimension to generate the video latent code $ z _ { 0 }$. In the forward diffusion procedure, Gaussian noise is gradually added to $ Z _ { 0 }$. Let $T$ be the total number of time steps in the diffusion process and $\left \{ \beta _ { t } \right \} _ { t=1 } ^ { T }$ the noise scheduler. The noisy latent code $ Z _ { t }$ at time step $t$ is expressed as follows:

\begin{equation}
	Z_t=\sqrt{\bar{\alpha}_t} Z_0+\sqrt{1-\bar{\alpha}_t} \epsilon, \epsilon \sim \mathcal{N}(0, I),
\end{equation}

\noindent
where $\bar{\alpha}_t=\prod_{i=1}^t \alpha_t, \alpha_t=1-\beta_t$ and $\epsilon$ is the adding noise code sampled from standard normal distribution $\mathcal{N}(0, I)$. As a result, the model aims to predict the noise $\epsilon$ from $Z_t$ and $t$. If the parameters of the 3D-UNet are denoted as $\theta$, the final objective function $\mathcal{L}$ is expressed as follows:

\begin{equation}
	\label{eq:1}
	\mathcal{L}=\mathbb{E}_{\epsilon \sim \mathcal{N}(0, I), t}\left[\left\|\epsilon-\epsilon_\theta\left(z_t, t\right)\right\|^2\right].
\end{equation}

\noindent
In the inference stage, if the video latent code is denoted as $Z_0$, noise is iteratively removed at each time step using the trained 3D-UNet. Finally, the video frames are reconstructed by the 2D-VAE decoder $D(\cdot)$.

\begin{figure*}[t]
	\centering
	\includegraphics[width=\linewidth]{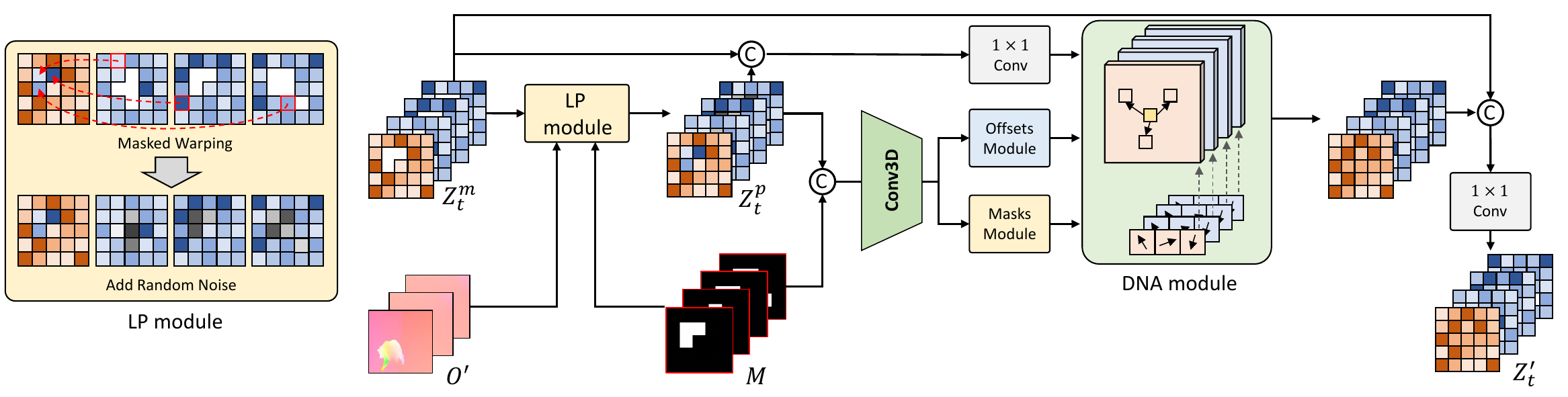}
	\caption{The structure of the proposed FFF module. First, the FFF module propagates the latent noise information from each frame to the first frame's latent noise to fill the masked areas. Next, deformable convolution is applied to reconstruct the latent-level distortions and structural information.}
	\label{fig:3mae}
\end{figure*}

\section{Proposed Approach}
\noindent
\textbf{FFF-VDI Trainig Stage.} Figure~\ref{fig:main} (a) shows the overall architecture of the proposed FFF-VDI. The FFF-VDI aims to generate new completed video frames $I^{\prime}=$ $\left\{I_1^{\prime}, I_2^{\prime}, \ldots, I_S^{\prime}\right\}$ using the masked $S$ input frames $I^m=\left\{I_1^m, I_2^m, \ldots, I_S^m\right\}$. In the FFF-VDI training stage, we pass each frame of $I$ through the 2D-VAE encoder $E$ to generate the video latent code $L_0 \in \mathbb{R} ^ {S \times C\times H \times W}$ and add time step noise to obtain the noisy latent code $L_t$. Then, we pass only the unmasked parts of the randomly masked frames $I^m$ through $E$ to generate the masked conditional latent code $L^m$. In other words, if the mask frames downsampled to the latent size are denoted as $M=\left\{M_1, M_2, \ldots, M_S\right\}$, the conditional latent code of the $i$-th frame $L_i^m$ is expressed as $L_i^m=E\left(I_i^m\right) \odot \left(1-M_i\right)$. Next, following the structure of Video LDM~\cite{rombach2022high}, we proceed with the process of merging the conditional latent and the noisy latent. As shown in Figure~\ref{fig:main}, the masked conditional latent $L^m$ is concatenated with $L_t$ along the channel dimension and then merged into the masked noisy latent code $Z_t^m$ through a $1 \times 1$ convolution layer.

Unlike existing methods that use pixel-level flow propagation for all frames, FFF-VDI applies noise latent level optical flow propagation only in the direction toward the first frame. To achieve this, we predict the masked optical flow $O^m$ from $I^m$ as shown in Figure~\ref{fig:main} and apply a flow completion module to convert it into the completed optical flow $O^{\prime}$. The flow completion module is widely used in traditional pixel propagation methods~\cite{zhang2022flow, kang2022error, li2022towards, zhang2022inertia, zhou2023propainter}, and we ultimately use the pretrained flow completion module proposed by ProPainter~\cite{zhou2023propainter}. $O^{\prime}$ and $Z_t^m$ are used as inputs to the First Frame Filling (FFF) module to fill the first frame masked noise latent $Z_{t_1}^m$. More specifically, the noise latent codes $Z_{t_{2: S}}^m$ from all frames except the first are propagated to $Z_{t_1}^m$ based on the optical flow. As a result, the FFF module generates the completed first frame noise latent $Z_t^{\prime}$.

Finally, $Z_t^{\prime}$ is used as the input to the pre-trained 3D-UNet for the denoising process. The objective function of FFF-VDI is the same as in Eq (\ref{eq:1}). We use the pre-trained image-to-video 3D-UNet from the Stable Video Diffusion~\cite{blattmann2023stable}. Therefore, to retrain the model for the video inpainting task, we fine-tune some layers of the 3D-UNet, as shown in Figure~\ref{fig:main}

\noindent
\textbf{FFF-VDI Testing Stage.} In Figure~\ref{fig:main} (b), we illustrate the testing stage of the proposed FFF-VDI, which is the diffusion inference process. During the generating stage, most video diffusion models use randomly initialized Gaussian noise, drawn from a normal distribution $N$, as the initial conditional latent. However, in the denoising process, when there is no text guidance, unwanted objects may appear in the inpainting task due to hallucination effects. To address this issue, instead of providing random noises for the entire areas during the denoising process, we use controlled noises to prevent the generation of unwanted objects by fully utilizing the given pixels. Specifically, we generate the original noisy latent from the masked video using DDIM inversion and propagate it to the first frame latent through the FFF module. 

\begin{table*}[t]
	\small
	\centering 
	\begin{tabular}{l|c|cccc|cccc}
		\hline
		\multirow{2}{*}{Methods} & \multirow{2}{*}{Publication} & \multicolumn{4}{c|}{YouTube-VOS} & \multicolumn{4}{c}{DAVIS}      \\ \cline{3-10} 
		&                        & PSNR$\uparrow$   & SSIM$\uparrow$    & VFID$\downarrow$  & $E_{warp}^*\downarrow$  & PSNR$\uparrow$  & SSIM$\uparrow$   & VFID$\downarrow$  & $E_{warp}^*\downarrow$  \\ \hline
		FGVC~\cite{gao2020flow}  & ECCV 2020                    & 29.67  & 0.9403  & 0.064 & 1.163 & 30.80  & 0.9497 & 0.165 & 1.571 \\
		STTN~\cite{zeng2020learning} & ECCV 2020                    & 32.34  & 0.9655  & 0.053 & 1.061 & 30.61 & 0.9560  & 0.149 & 1.438 \\
		TSAM~\cite{zou2021progressive} & CVPR 2021                    & 30.22  & 0.9468  & 0.070  & 1.014 & 30.67 & 0.9548 & 0.146 & 1.235 \\
		FuseFormer~\cite{liu2021fuseformer} & ICCV 2021                    & 33.32  & 0.9681  & 0.053 & 1.053 & 32.59 & 0.9701 & 0.137 & 1.349 \\
		ISVI~\cite{zhang2022inertia} & CVPR 2022                    & 30.34  & 0.9458  & 0.077 & 1.008 & 32.17 & 0.9588 & 0.189 & 1.291 \\
		FGT~\cite{zhang2022flow} & ECCV 2022                    & 32.17  & 0.9599  & 0.054 & 1.025 & 32.86 & 0.9650  & 0.129 & 1.323 \\
		E$^2$FGVI~\cite{li2022towards} & CVPR 2022                    & 33.71  & 0.9700    & 0.046 & 1.013 & 33.01 & 0.9721 & 0.116 & 1.289 \\
		ProPainter~\cite{zhou2023propainter} & ICCV 2023                    & 34.43  & 0.9735  & 0.042 & 0.974 & 34.47 & 0.9776 & 0.098 & 1.187 \\ \hline\hline
		FFF-VDI (Ours)  &       & \textbf{35.06} & \textbf{0.9812} & \textbf{0.031} & \textbf{0.937} & \textbf{35.03} & \textbf{0.9834} & \textbf{0.075} & \textbf{1.009} \\ \hline
	\end{tabular}
	\caption{Quantitative comparison on the YouTube-VOS and DAVIS datasets. $E_{warp}^*$ denotes $E_{warp}\left(\times 10^{-3}\right)$.}
	\label{Table:results}
\end{table*}

To explain in more detail, we first pass the masked video through a 2D-VAE encoder to generate the masked conditional latent code $L^m$. Next, using the DDIM inversion process, we transform $L^m$ into the inverted noise latent code $L_T^i$ at time step $T$. Then, we merge different noise $N \sim \mathcal{N}(0, I)$ and $L_T^i$ with $L^m$ through the same convolution layer to generate $Z_T$ and $Z_T^i$, respectively. Since FFF-VDI completes the first frame latent code using only the latent information from other frames, we use the newly concatenated latent code, formed by concatenating $Z_{T_1}$ and $Z_{T_{2:S}}^i$ along the temporal axis, as the input for FFF module. Finally, we convert the latent code that has passed through the FFF module into the final denoised latent code using the standard diffusion denoising process. Similar to existing video inpainting evaluation processes, we composite the generated masked areas with the masked original video to produce the final video.

\noindent
\textbf{First Frame Filling Module.} Figure~\ref{fig:3mae} shows the structure of the proposed FFF module, which includes the Latent Propagation (LP) module and the Deformable Noise Alignment (DNA) module, First, LP module use the downsampled optical flow $O^{\prime}$ and the downsampled mask map $M$ to warp the noisy latent information $Z_{t_{2:S}}^m$ from other frames to the first frame $Z_{t_1}^m$. To preserve the original first frame information, the latent information is propagated only to the masked regions of $Z_{t_1}^m$, which correspond to the unmasked regions of $Z_{t_{2:S}}^m$. Therefore, the warping process can be expressed as follows:

\begin{equation}
	Z_{t_1}^p=\left(1-M_1\right)Z_{t_1}^m+M_1\sum_{i=2}^S\mathcal{W}\left(\left(1-M_i\right)Z_{t_i}^w, O_{i \rightarrow 1}^{\prime}\right),
\end{equation}

\noindent
where $\mathcal{W}(\cdot)$ denotes warping operation. Additionally, we sequentially apply optical flow to warp the latent from the $i$-th frame to the first frame. In other words, the flow $O_{i \rightarrow 1^{\prime}}^{\prime}$, which warps from the $i$-th frame to the first frame, is composed of the sequential set $\left\{O_{i \rightarrow i-1}^{\prime}, O_{i-1 \rightarrow i-2}^{\prime}, \ldots, O_{2 \rightarrow 1}^{\prime}\right\} $. Next, we add random noise to the masked regions of $Z_{t_{2:S}}^m$ and combine it with the first frame noisy latent to generate the propagated $Z_t^p$. This process can be expressed as follows:

\begin{equation}
	Z_t^p=Z_1^p \| \left(Z_{t_{2: S}}^m+\epsilon M_{2:S}\right), \epsilon \sim \mathcal{N}\left(0, I\right),
\end{equation}

\noindent
where $\|$ is the concatenation operation along the temporal axis. 

Unlike the first frame where flow propagation is applied, we fill the masked areas of future frames with random noise, which can lead to a decrease in temporal consistency during the diffusion denoising process. Therefore, we add a noise refinement process to maintain the temporal consistency of the generated noisy latent. For this, we introduce a DNA process using a deformable convolution network (DCN). Although a similar approach is proposed in E2FGVI~\cite{li2022towards}, our model differs in that it does not use optical flow for DCN offset prediction. This minimizes the distortion and consistency errors in temporal noisy latents caused by incorrect flow completion. Instead, as shown in Figure~\ref{fig:3mae}, the DNA module applies 3D convolution layers to directly learn DCN offsets from the noisy latent. As a result, we combine the refined noisy latent with $Z_t^m$ to generate the final noisy latent $Z_t^{\prime}$. Specifically, we first concatenate the previously generated $Z_t^p$ and the mask $M$ and pass them through a stack of 3D convolution layers. We then separate this feature to generate the offset $o$ and the DCN modulation mask $m$. $m$ adjusts the contribution of each noisy latent pixel in the DCN. Our DNA process is expressed as follows:

\begin{equation}
	\hat{Z}_{t_s}=\mathcal{R}\left(\mathcal{D}\left(\hat{Z}_{t_{s+1}}, o_{s \rightarrow s+1}, m_{s \rightarrow s+1}\right), \hat{Z}_{t_s}\right),	
\end{equation}

\noindent
where $\mathcal{D}(\cdot)$ denotes deformable convolution, and $\mathcal{R}(\cdot)$ denotes the convolution layers that fuse the aligned and current features. Additionally, $\hat{Z}_t$ is the feature where $Z_t^m$ and $Z_t^p$ are merged. As a result, we combine the refined noisy latent with $Z_t^m$ to generate the final noisy latent $Z_t^{\prime}$.

\section{Experiments}
\noindent
\textbf{Datasets.} To fairly compare previous state-of-the-art models with the proposed FFF-VDI, we use the YouTube-VOS~\cite{xu2018youtube} training set as the training data. Additionally, for model evaluation, we use the widely known YouTube-VOS~\cite{xu2018youtube} and DAVIS~\cite{perazzi2016benchmark} test sets as evaluation datasets. The YouTube-VOS test set consists of 508 video clips, and the DAVIS test set consists of 90 video clips. For the DAVIS test set, we follow the approach of ProPainter~\cite{zhou2023propainter} and E2FGVI~\cite{li2022towards}, using 50 video clips for evaluation.

\noindent
\textbf{Training Details and Evaluation Metrics.} In this paper, we use the optical flow prediction model RAFT~\cite{teed2020raft} to extract optical flow. Additionally, we use the pre-trained flow completion model proposed by ProPainter~\cite{zhou2023propainter} to inpaint the masked optical flow. To ensure a fair comparison with prior studies, all videos are resized to $432 \times 240$. However, due to the specific input size required by the 3D-UNet of the Video LDM~\cite{rombach2022high}, we pad the input frames to change its size to $448 \times 256$. Also, we follow previous methods~\cite{li2022towards, liu2021fuseformer, zhou2023propainter} to randomly generate stationary and object masks to simulate masks for video completion and object removal tasks. Our pre-trained image-to-video diffusion model follows the official implementation of Stable Video Diffusion~\cite{blattmann2023stable}, and we fine-tune the model by freezing all layers of the 3D-UNet except for the temporal transformer block. In this paper, we set the batch size to 4, the initial learning rate to $10^{-5}$, and train for a total of 100,000 iterations with Adam~\cite{kingma2014adam} optimizer. Our method is implemented using the PyTorch framework and trained on four NVIDIA RTX A6000 GPUs.

\begin{figure*}[t]
	\centering
	\includegraphics[width=\linewidth]{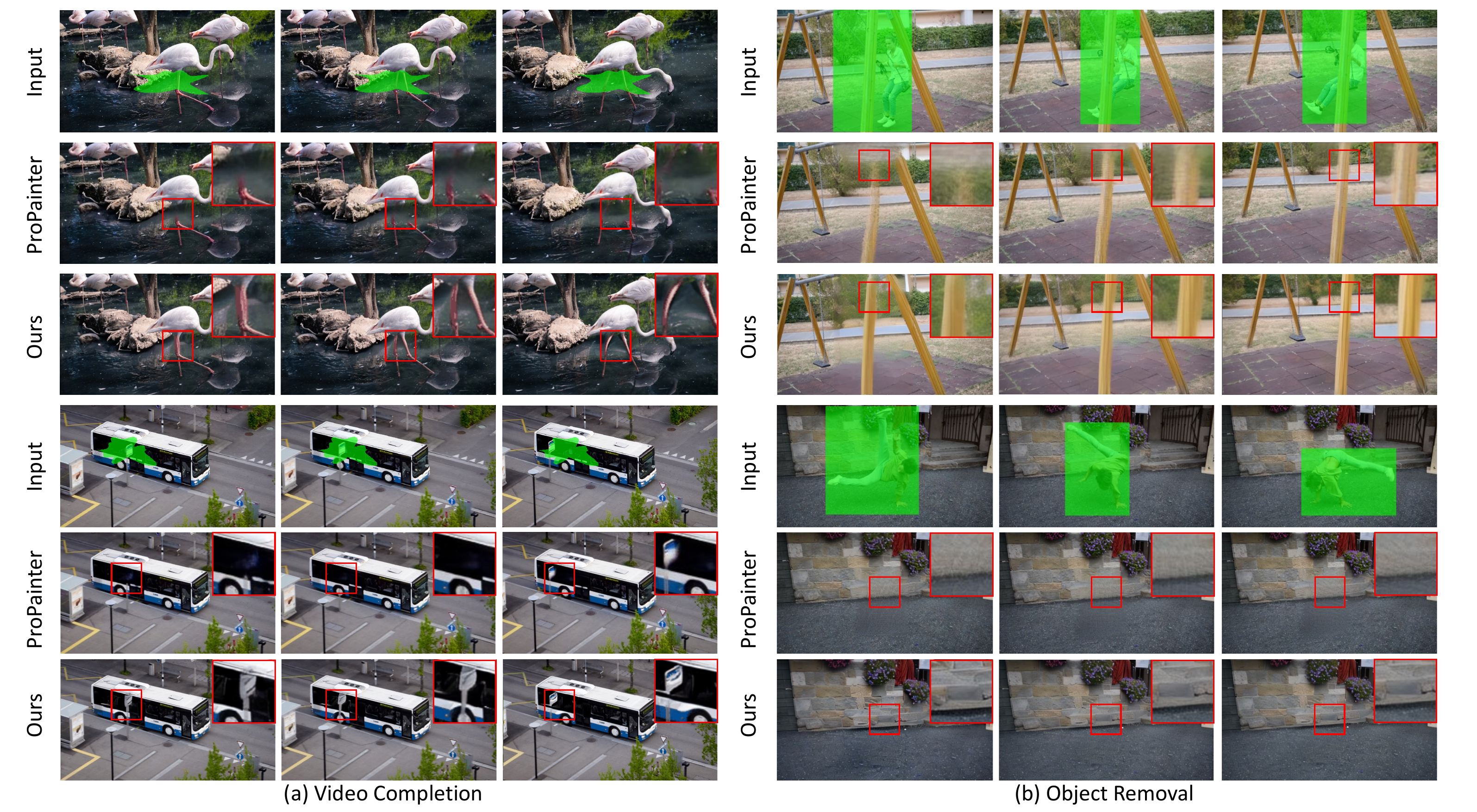}
	\caption{Qualitative comparisons on both video completion and object removal. The proposed FFF-VDI demonstrates robust video inpainting performance in masked areas compared to the existing flow propagation-based model, ProPainter.}
	\label{fig:result}
\end{figure*}

\begin{figure}[t]
	\centering
	\includegraphics[width=\linewidth]{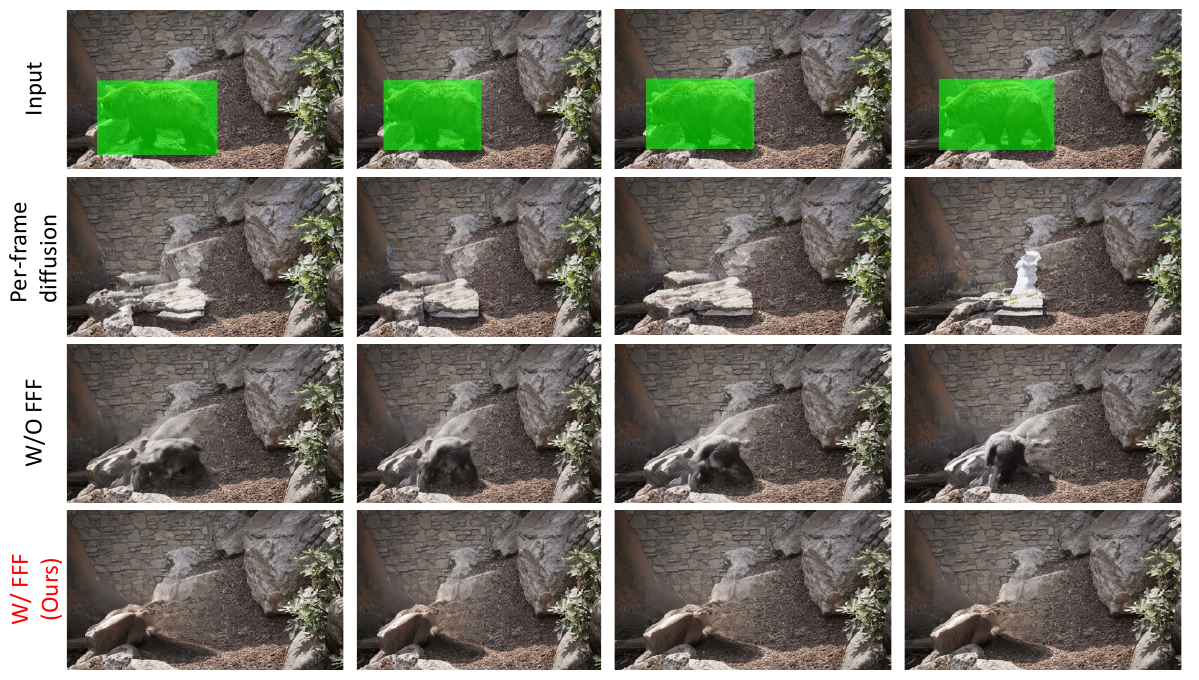}
	\caption{Visualization results with and without the proposed FFF module. Per-frame diffusion is the result of inpainting each frame with stable diffusion.}
	\label{fig:recon}
\end{figure}

\begin{figure}[t]
	\centering
	\includegraphics[width=\linewidth]{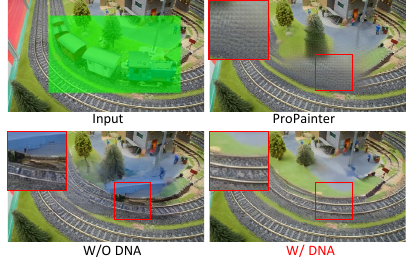}
	\caption{Visualization results with and without the DNA module.}
	\label{fig:comp}
\end{figure}

For model evaluation, we use PSNR and SSIM~\cite{wang2004image}, which are traditionally used in video inpainting tasks. Additionally, to assess the temporal consistency and smoothness of the resulting video sequences, we use the flow warping error $E_{warp}$~\cite{lai2018learning} as an evaluation metric. Furthermore, following recent video inpainting researches~\cite{liu2021fuseformer, li2022towards, zhou2023propainter}, we report the VFID~\cite{wang2018video} score to measure the perceptual similarity between the output and ground truth videos.

\subsection{Results}
\noindent
\textbf{Quantitative results.} In Table~\ref{Table:results}, we quantitatively compare our proposed FFF-VDI with state-of-the-art methods on the YouTube-VOS and DAVIS datasets. The removal mask maps for testing are the same as the official test mask maps provided by ProPainter~\cite{zhou2023propainter}. As shown in the table, our method significantly outperforms all existing methods in the video completion task. Notably, we demonstrate greater performance improvement on the perceptual metric VFID than on traditional metrics like PSNR and SSIM. This is because traditional metrics tend to give higher scores to blurry textures rather than natural textures. Therefore, traditional metrics fail to properly evaluate the generation quality of blurry images resulting from incorrect optical flow predictions in previous flow propagation-based methods. Furthermore, flow propagation-based methods typically demonstrate high time consistency and excellent $E_{warp}$ performance because they use flow to predict the next frame. However, due to their limited generative capability, they cannot effectively reconstruct missing pixel information. Therefore, the results in Table~\ref{Table:results} show that the proposed FFF-VDI can effectively integrate the superior generative ability and time consistency of the image-to-video diffusion model into the video inpainting task without direct pixel propagation.

\noindent
\textbf{Qualitative results.} Figure~\ref{fig:result} shows the qualitative evaluation results of the previous state-of-the-art method, ProPainter~\cite{zhou2023propainter}, and the proposed FFF-VDI across various scenarios. For the sake of a fair comparison, both ProPainter and FFF-VDI use the same number of reference frames. As mentioned earlier, visual quality is often not adequately reflected in quantitative metrics, making it crucial to compare qualitative evaluations. Therefore, in Figure 4, we present (a) random mask video inpainting results and (b) rough rectangular mask object removal results. As illustrated, the proposed model demonstrates significantly higher performance in propagating actual contextual information to the occluded areas and generating new content that cannot be observed, compared to existing methods. Particularly, the proposed method achieves natural video synthesis with high time consistency and fewer distortions in areas where flow propagation is not possible, due to its lower dependence on optical flow.

\begin{figure*}[t]
	\centering
	\includegraphics[width=\linewidth]{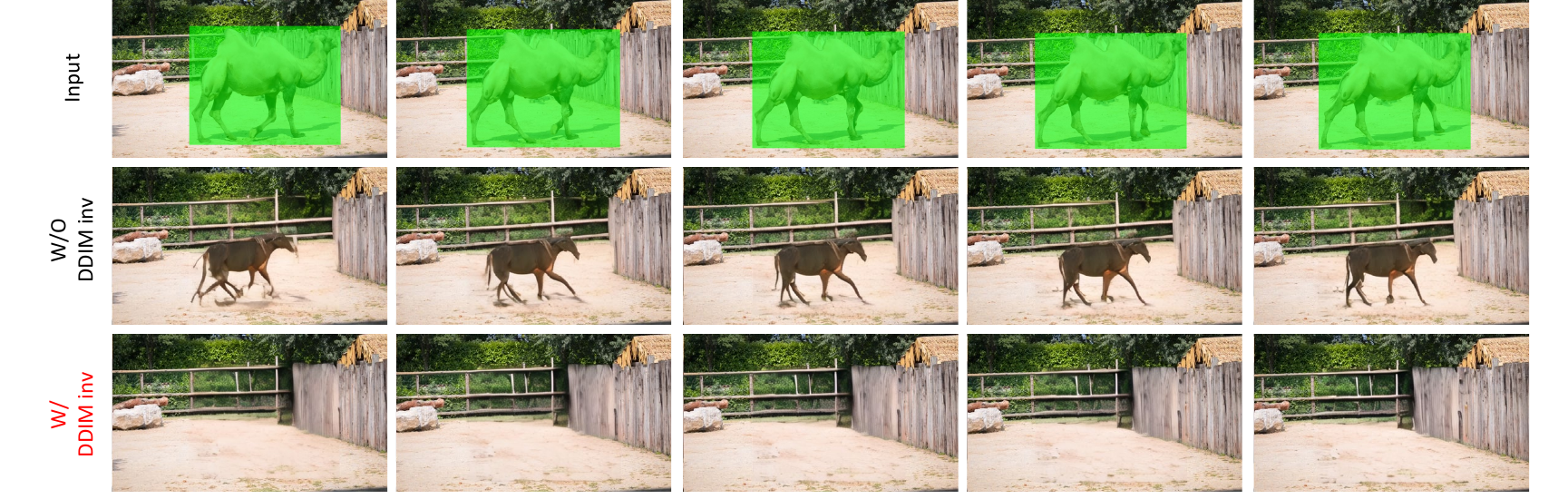}
	\caption{Comparison of generated images with and without DDIM inversion.}
	\label{fig:inv}
\end{figure*}

\subsection{Ablation Analysis}
\label{ablation}

\begin{table*}[t]
	\small
	\centering 
	\begin{tabular}{cccc|cccc|cccc}
		\hline
		\multicolumn{4}{c|}{Method}                                                                                 & \multicolumn{4}{c|}{YouTube-VOS}                                                           & \multicolumn{4}{c}{DAVIS}                                                                  \\ \hline
		\multicolumn{1}{c|}{\multirow{2}{*}{Per-frame}} & \multicolumn{2}{c|}{FFF-VDI}  & \multirow{2}{*}{DDIM inv} & \multirow{2}{*}{PSNR$\uparrow$} & \multirow{2}{*}{SSIM$\uparrow$} & \multirow{2}{*}{VFID$\downarrow$} & \multirow{2}{*}{$E_{warp}^*\downarrow$} & \multirow{2}{*}{PSNR$\uparrow$} & \multirow{2}{*}{SSIM$\uparrow$} & \multirow{2}{*}{VFID$\downarrow$} & \multirow{2}{*}{$E_{warp}^*\downarrow$} \\ \cline{2-3}
		\multicolumn{1}{c|}{} & LP & \multicolumn{1}{c|}{DNA} & & & & & & & & & \\ \hline
		\ding{51} & & & & 32.21 & 0.9532 & 0.063 & 1.592 & 30.59 & 0.9572 & 0.128 & 1.478 \\
		& \ding{51} & & & 34.03 & 0.9783 & 0.043 & 1.142 & 34.11 & 0.9798 & 0.089 & 1.206 \\
		& \ding{51} & \ding{51} & & 35.01 & 0.9798 & 0.031 & 0.940 & 35.10 & 0.9814 & 0.074 & 1.034 \\
		& \ding{51} & \ding{51} & \ding{51} & 35.06 & 0.9812  & 0.031 & 0.937 & 35.03 & 0.9834 & 0.075 & 1.009 \\ \hline
	\end{tabular}
	\caption{Comparison of random mask video inpainting performance based on the proposed contribution combinations.}
	\label{Table:comp}
\end{table*}

\noindent
\textbf{Effect of first frame filling.} Table~\ref{Table:comp} and Figure~\ref{fig:recon} demonstrate the effectiveness of the proposed FFF module. First, the second row of Figure~\ref{fig:recon} shows the results of applying Stable Diffusion~\cite{rombach2022high} frame-by-frame to each masked area, where we perform inpainting with an empty text input. While the diffusion-based model provides natural fill-in images on a frame-by-frame basis due to its powerful generative capabilities, it exhibits very low temporal consistency quality. The third row shows our baseline model without the FFF module. In this case, even without applying latent propagation, the pre-trained video diffusion model's generative capabilities allow for the creation of a fairly temporally consistent video. However, the video generated with random noise does not reflect the actual context information of the initial frames. Therefore, as the mask moves and the real background behind the mask is revealed, perceptual inconsistencies arise between past and future frames. In other words, it is challenging for the existing diffusion model to model long-range temporal consistency, which is a major reason why applying video diffusion models to video inpainting tasks is difficult. However, the proposed FFF-VDI pre-completes the accurate background information of the first frame using the latent codes from the other frames, thus generating the most natural background video, as shown in the last row of Figure~\ref{fig:recon}. These results are also well-reflected in the quantitative analysis results in Table~\ref{Table:comp}.

\noindent
\textbf{Effect of deformable noise alignment.} Table~\ref{Table:comp} and Figure~\ref{fig:comp} demonstrate the effect of the DNA module. As shown in Figure~\ref{fig:comp}, when the mask size is very large, the reference frame provides limited visual cues, resulting in images with unnatural structures. Although ProPainter uses optical flow-based deformable convolution, it fails to reconstruct structural information when the flow information is also very limited, as seen in Figure~\ref{fig:comp}. The proposed deformable alignment minimizes distortion in the denoised result images at the noise latent level instead of relying on optical flow, aiding in the generation of natural images.

\noindent
\textbf{Effect of DDIM inversion.} Figure~\ref{fig:inv} compares the generated image results of FFF-VDI with and without the DDIM inversion process. As shown in the second row, due to the powerful generative capabilities of the pre-trained video diffusion model, hallucination effects occur, generating unwanted objects when the target mask size is large. However, as shown in the third row, when noise latent information is provided through DDIM inversion during the inference stage, FFF-VDI is able to generate normal videos. This trend is also reflected in the quantitative analysis results in Table~\ref{Table:comp}, although the effect of DDIM inversion is less significant due to the relatively smaller mask size used by ProPainter. In fact, most diffusion models use text guidance to address hallucination effects, but providing text guidance for backgrounds is challenging in video inpainting tasks. Therefore, the proposed FFF-VDI effectively resolves the issues of existing video diffusion models for video inpainting tasks and integrates the model effectively.


\section{Conclusion}
We introduce FFF-VDI, a novel video diffusion inpainting model based on First Frame Filling. Unlike traditional propagation-based models, FFF-VDI uses noise latent propagation for the first frame and leverages the temporal consistency of fine-tuned image-to-video diffusion models for subsequent frames. FFF-VDI addresses two major challenges in video inpainting. It ensures natural video generation when masks move by reconstructing the actual appearance behind the mask and minimizes object hallucination by using DDIM inversion. Comparative experiments show that FFF-VDI outperforms previous methods in video completion and object removal, especially with large and rough masks. This makes FFF-VDI a robust and effective solution for complex video inpainting tasks.

\section{Acknowledgement}
This work was supported by the National Research Foundation of Korea (NRF) grant funded by the Korea government (MSIT)(No. RS-2024-00423362) and (MSIT)(No. RS-2024-00340745). 

\bibliography{aaai25}

\begin{thebibliography}{24}
\providecommand{\natexlab}[1]{#1}

\bibitem[{Blattmann et~al.(2023)Blattmann, Dockhorn, Kulal, Mendelevitch,
  Kilian, Lorenz, Levi, English, Voleti, Letts et~al.}]{blattmann2023stable}
Blattmann, A.; Dockhorn, T.; Kulal, S.; Mendelevitch, D.; Kilian, M.; Lorenz,
  D.; Levi, Y.; English, Z.; Voleti, V.; Letts, A.; et~al. 2023.
\newblock Stable video diffusion: Scaling latent video diffusion models to
  large datasets.
\newblock \emph{arXiv preprint arXiv:2311.15127}.

\bibitem[{Gao et~al.(2020)Gao, Saraf, Huang, and Kopf}]{gao2020flow}
Gao, C.; Saraf, A.; Huang, J.-B.; and Kopf, J. 2020.
\newblock Flow-edge guided video completion.
\newblock In \emph{Computer Vision--ECCV 2020: 16th European Conference,
  Glasgow, UK, August 23--28, 2020, Proceedings, Part XII 16}, 713--729.
  Springer.

\bibitem[{Gu et~al.(2023)Gu, Yu, Fan, and Zhang}]{gu2023flow}
Gu, B.; Yu, Y.; Fan, H.; and Zhang, L. 2023.
\newblock Flow-guided diffusion for video inpainting.
\newblock \emph{arXiv preprint arXiv:2311.15368}.

\bibitem[{Ho et~al.(2022)Ho, Salimans, Gritsenko, Chan, Norouzi, and
  Fleet}]{ho2022video}
Ho, J.; Salimans, T.; Gritsenko, A.; Chan, W.; Norouzi, M.; and Fleet, D.~J.
  2022.
\newblock Video diffusion models.
\newblock \emph{Advances in Neural Information Processing Systems}, 35:
  8633--8646.

\bibitem[{Kang, Oh, and Kim(2022)}]{kang2022error}
Kang, J.; Oh, S.~W.; and Kim, S.~J. 2022.
\newblock Error compensation framework for flow-guided video inpainting.
\newblock In \emph{European conference on computer vision}, 375--390. Springer.

\bibitem[{Kingma and Ba(2014)}]{kingma2014adam}
Kingma, D.~P.; and Ba, J. 2014.
\newblock Adam: A method for stochastic optimization.
\newblock \emph{arXiv preprint arXiv:1412.6980}.

\bibitem[{Kingma and Welling(2013)}]{kingma2013auto}
Kingma, D.~P.; and Welling, M. 2013.
\newblock Auto-encoding variational bayes.
\newblock \emph{arXiv preprint arXiv:1312.6114}.

\bibitem[{Lai et~al.(2018)Lai, Huang, Wang, Shechtman, Yumer, and
  Yang}]{lai2018learning}
Lai, W.-S.; Huang, J.-B.; Wang, O.; Shechtman, E.; Yumer, E.; and Yang, M.-H.
  2018.
\newblock Learning blind video temporal consistency.
\newblock In \emph{Proceedings of the European conference on computer vision
  (ECCV)}, 170--185.

\bibitem[{Li et~al.(2022)Li, Lu, Qin, Guo, and Cheng}]{li2022towards}
Li, Z.; Lu, C.-Z.; Qin, J.; Guo, C.-L.; and Cheng, M.-M. 2022.
\newblock Towards an end-to-end framework for flow-guided video inpainting.
\newblock In \emph{Proceedings of the IEEE/CVF conference on computer vision
  and pattern recognition}, 17562--17571.

\bibitem[{Liu et~al.(2021)Liu, Deng, Huang, Shi, Lu, Sun, Wang, Dai, and
  Li}]{liu2021fuseformer}
Liu, R.; Deng, H.; Huang, Y.; Shi, X.; Lu, L.; Sun, W.; Wang, X.; Dai, J.; and
  Li, H. 2021.
\newblock Fuseformer: Fusing fine-grained information in transformers for video
  inpainting.
\newblock In \emph{Proceedings of the IEEE/CVF international conference on
  computer vision}, 14040--14049.

\bibitem[{Mokady et~al.(2023)Mokady, Hertz, Aberman, Pritch, and
  Cohen-Or}]{mokady2023null}
Mokady, R.; Hertz, A.; Aberman, K.; Pritch, Y.; and Cohen-Or, D. 2023.
\newblock Null-text inversion for editing real images using guided diffusion
  models.
\newblock In \emph{Proceedings of the IEEE/CVF Conference on Computer Vision
  and Pattern Recognition}, 6038--6047.

\bibitem[{Perazzi et~al.(2016)Perazzi, Pont-Tuset, McWilliams, Van~Gool, Gross,
  and Sorkine-Hornung}]{perazzi2016benchmark}
Perazzi, F.; Pont-Tuset, J.; McWilliams, B.; Van~Gool, L.; Gross, M.; and
  Sorkine-Hornung, A. 2016.
\newblock A benchmark dataset and evaluation methodology for video object
  segmentation.
\newblock In \emph{Proceedings of the IEEE conference on computer vision and
  pattern recognition}, 724--732.

\bibitem[{Rombach et~al.(2022)Rombach, Blattmann, Lorenz, Esser, and
  Ommer}]{rombach2022high}
Rombach, R.; Blattmann, A.; Lorenz, D.; Esser, P.; and Ommer, B. 2022.
\newblock High-resolution image synthesis with latent diffusion models.
\newblock In \emph{Proceedings of the IEEE/CVF conference on computer vision
  and pattern recognition}, 10684--10695.

\bibitem[{Teed and Deng(2020)}]{teed2020raft}
Teed, Z.; and Deng, J. 2020.
\newblock Raft: Recurrent all-pairs field transforms for optical flow.
\newblock In \emph{Computer Vision--ECCV 2020: 16th European Conference,
  Glasgow, UK, August 23--28, 2020, Proceedings, Part II 16}, 402--419.
  Springer.

\bibitem[{Wang et~al.(2018)Wang, Liu, Zhu, Liu, Tao, Kautz, and
  Catanzaro}]{wang2018video}
Wang, T.-C.; Liu, M.-Y.; Zhu, J.-Y.; Liu, G.; Tao, A.; Kautz, J.; and
  Catanzaro, B. 2018.
\newblock Video-to-video synthesis.
\newblock \emph{arXiv preprint arXiv:1808.06601}.

\bibitem[{Wang et~al.(2004)Wang, Bovik, Sheikh, and Simoncelli}]{wang2004image}
Wang, Z.; Bovik, A.~C.; Sheikh, H.~R.; and Simoncelli, E.~P. 2004.
\newblock Image quality assessment: from error visibility to structural
  similarity.
\newblock \emph{IEEE transactions on image processing}, 13(4): 600--612.

\bibitem[{Xu et~al.(2018)Xu, Yang, Fan, Yang, Yue, Liang, Price, Cohen, and
  Huang}]{xu2018youtube}
Xu, N.; Yang, L.; Fan, Y.; Yang, J.; Yue, D.; Liang, Y.; Price, B.; Cohen, S.;
  and Huang, T. 2018.
\newblock Youtube-vos: Sequence-to-sequence video object segmentation.
\newblock In \emph{Proceedings of the European conference on computer vision
  (ECCV)}, 585--601.

\bibitem[{Zeng, Fu, and Chao(2020)}]{zeng2020learning}
Zeng, Y.; Fu, J.; and Chao, H. 2020.
\newblock Learning joint spatial-temporal transformations for video inpainting.
\newblock In \emph{Computer Vision--ECCV 2020: 16th European Conference,
  Glasgow, UK, August 23--28, 2020, Proceedings, Part XVI 16}, 528--543.
  Springer.

\bibitem[{Zhang, Fu, and Liu(2022{\natexlab{a}})}]{zhang2022flow}
Zhang, K.; Fu, J.; and Liu, D. 2022{\natexlab{a}}.
\newblock Flow-guided transformer for video inpainting.
\newblock In \emph{European Conference on Computer Vision}, 74--90. Springer.

\bibitem[{Zhang, Fu, and Liu(2022{\natexlab{b}})}]{zhang2022inertia}
Zhang, K.; Fu, J.; and Liu, D. 2022{\natexlab{b}}.
\newblock Inertia-guided flow completion and style fusion for video inpainting.
\newblock In \emph{Proceedings of the IEEE/CVF conference on computer vision
  and pattern recognition}, 5982--5991.

\bibitem[{Zhang et~al.(2024)Zhang, Wu, Wang, Luo, Zhang, Zhao, Vajda, Metaxas,
  and Yu}]{zhang2024avid}
Zhang, Z.; Wu, B.; Wang, X.; Luo, Y.; Zhang, L.; Zhao, Y.; Vajda, P.; Metaxas,
  D.; and Yu, L. 2024.
\newblock AVID: Any-Length Video Inpainting with Diffusion Model.
\newblock In \emph{Proceedings of the IEEE/CVF Conference on Computer Vision
  and Pattern Recognition}, 7162--7172.

\bibitem[{Zhou et~al.(2023)Zhou, Li, Chan, and Loy}]{zhou2023propainter}
Zhou, S.; Li, C.; Chan, K.~C.; and Loy, C.~C. 2023.
\newblock Propainter: Improving propagation and transformer for video
  inpainting.
\newblock In \emph{Proceedings of the IEEE/CVF International Conference on
  Computer Vision}, 10477--10486.

\bibitem[{Zi et~al.(2024)Zi, Zhao, Qi, Wang, Shi, Chen, Liang, Wong, and
  Zhang}]{zi2024cococo}
Zi, B.; Zhao, S.; Qi, X.; Wang, J.; Shi, Y.; Chen, Q.; Liang, B.; Wong, K.-F.;
  and Zhang, L. 2024.
\newblock CoCoCo: Improving Text-Guided Video Inpainting for Better
  Consistency, Controllability and Compatibility.
\newblock \emph{arXiv preprint arXiv:2403.12035}.

\bibitem[{Zou et~al.(2021)Zou, Yang, Liu, and Lee}]{zou2021progressive}
Zou, X.; Yang, L.; Liu, D.; and Lee, Y.~J. 2021.
\newblock Progressive temporal feature alignment network for video inpainting.
\newblock In \emph{Proceedings of the IEEE/CVF Conference on Computer Vision
  and Pattern Recognition}, 16448--16457.

\end{thebibliography}

\end{document}